\documentclass[letterpaper]{article} % DO NOT CHANGE THIS
\usepackage{aaai24}  % DO NOT CHANGE THIS
\usepackage{times}  % DO NOT CHANGE THIS
\usepackage{helvet}  % DO NOT CHANGE THIS
\usepackage{courier}  % DO NOT CHANGE THIS
\usepackage[hyphens]{url}  % DO NOT CHANGE THIS
\usepackage{graphicx} % DO NOT CHANGE THIS
\usepackage{natbib}  % DO NOT CHANGE THIS AND DO NOT ADD ANY OPTIONS TO IT
\usepackage{caption} % DO NOT CHANGE THIS AND DO NOT ADD ANY OPTIONS TO IT
\usepackage{algorithm}
\usepackage{algorithmic}

\usepackage{newfloat}
\usepackage{listings}
\DeclareCaptionStyle{ruled}{labelfont=normalfont,labelsep=colon,strut=off} % DO NOT CHANGE THIS
\floatstyle{ruled}
\newfloat{listing}{tb}{lst}{}
\floatname{listing}{Listing}
\usepackage{amsmath}
\usepackage{amsfonts}
\usepackage{amssymb}
\DeclareMathOperator*{\argmax}{arg\,max}

\title{The Role of Higher-Order Cognitive Models in Active Learning}
\author {
    Oskar Keurulainen, %\textsuperscript{\rm 1},
    Gokhan Alcan, %\textsuperscript{\rm 1},
    Ville Kyrki%\textsuperscript{\rm 1}
}
\usepackage{bibentry}
\begin{document}

\maketitle

\begin{abstract}

Building machines capable of efficiently collaborating with humans has been a longstanding goal in artificial intelligence. Especially in the presence of uncertainties, optimal cooperation often requires that humans and artificial agents model each other's behavior and use these models to infer underlying goals, beliefs or intentions, potentially involving multiple levels of recursion. Empirical evidence for such higher-order cognition in human behavior is also provided by previous works in cognitive science, linguistics, and robotics. We advocate for a new paradigm for active learning for human feedback that utilises humans as active data sources while accounting for their higher levels of agency. In particular, we discuss how increasing level of agency results in qualitatively different forms of rational communication between an active learning system and a teacher. Additionally, we provide a practical example of active learning using a higher-order cognitive model. This is accompanied by a computational study that underscores the unique behaviors that this model produces.

\end{abstract}

\insert\footins{\noindent\footnotesize This work was financially supported by the Academy of Finland grant numbers 345661 and 347199. \\ All authors are with the Intelligent Robotics Group, Department of Electrical Engineering and Automation, Aalto University, Finland. \\ E-mails: \texttt{firstname.lastname@aalto.fi}.}

\section{Introduction}

The active involvement of humans in training of Artificial Intelligence (AI) systems is becoming increasingly common. Compared to passive data sources, a key advantage of learning from human feedback is that humans can be queried by active learning methods, thereby maximising the usefulness of the expected data. However, a computational model of human behavior, a necessary component of such AI systems, motivates multidisciplinary research efforts to better understand human-AI interactions. A typical modeling assumption in learning from human feedback is Boltzmann rationality, which models goal-directed human behavior as approximately rational with respect to the task, but does not account for potential human situational awareness that extends beyond the task itself.

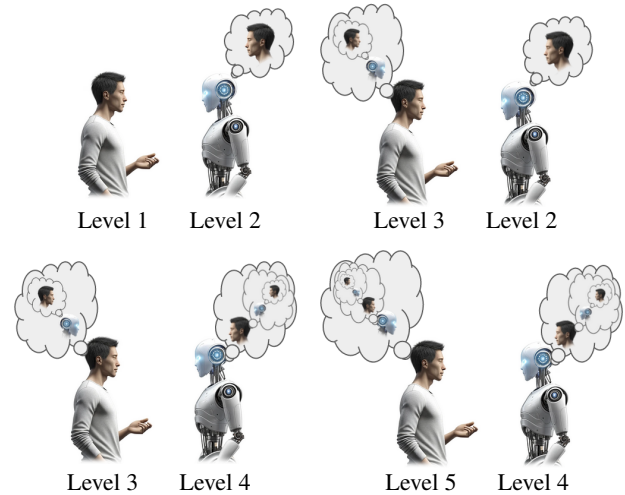
\begin{figure}[t!]
\vspace{0.2cm}
	\centering
	\def\svgwidth{1\linewidth}
	{
 \fontsize{9}{9}
		%% Creator: Inkscape 1.3.2 (1:1.3.2+202311252150+091e20ef0f), www.inkscape.org
%% Accompanies image file 'fig1.pdf' (pdf, eps, ps)
\begingroup%
  \makeatletter%
  \providecommand\color[2][]{%
    \errmessage{(Inkscape) Color is used for the text in Inkscape, but the package 'color.sty' is not loaded}%
    \renewcommand\color[2][]{}%
  }%
  \providecommand\transparent[1]{%
    \errmessage{(Inkscape) Transparency is used (non-zero) for the text in Inkscape, but the package 'transparent.sty' is not loaded}%
    \renewcommand\transparent[1]{}%
  }%
  \providecommand\rotatebox[2]{#2}%
  \newcommand*\fsize{\dimexpr\f@size pt\relax}%
  \newcommand*\lineheight[1]{\fontsize{\fsize}{#1\fsize}\selectfont}%
  \ifx\svgwidth\undefined%
    \setlength{\unitlength}{1065.94926489bp}%
    \ifx\svgscale\undefined%
      \relax%
    \else%
      \setlength{\unitlength}{\unitlength * \real{\svgscale}}%
    \fi%
  \else%
    \setlength{\unitlength}{\svgwidth}%
  \fi%
  \global\let\svgwidth\undefined%
  \global\let\svgscale\undefined%
  \makeatother%
  \begin{picture}(1,0.81563471)%
    \lineheight{1}%
    \setlength\tabcolsep{0pt}%
    \put(0,0){\includegraphics[width=\unitlength,page=1]{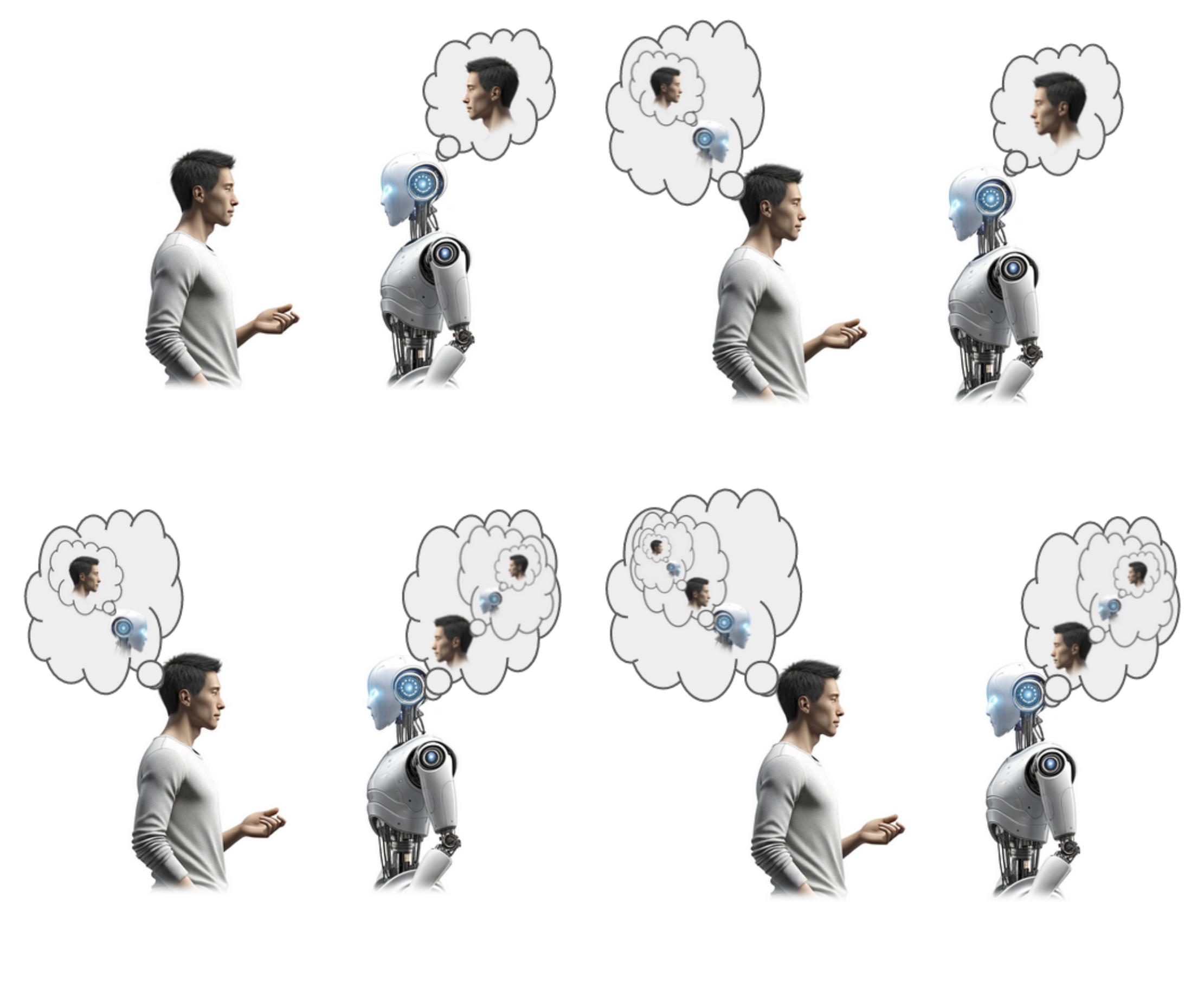}}%
    \put(0.11992539,0.44445321){\makebox(0,0)[lt]{\lineheight{1.25}\smash{\begin{tabular}[t]{l}$\text{Level 1}$\end{tabular}}}}%
    \put(0.29731067,0.44445321){\makebox(0,0)[lt]{\lineheight{1.25}\smash{\begin{tabular}[t]{l}$\text{Level 2}$\end{tabular}}}}%
    \put(0.58801912,0.44445321){\makebox(0,0)[lt]{\lineheight{1.25}\smash{\begin{tabular}[t]{l}$\text{Level 3}$\end{tabular}}}}%
    \put(0.76540248,0.44445321){\makebox(0,0)[lt]{\lineheight{1.25}\smash{\begin{tabular}[t]{l}$\text{Level 2}$\end{tabular}}}}%
    \put(0.10338981,0.03164189){\makebox(0,0)[lt]{\lineheight{1.25}\smash{\begin{tabular}[t]{l}$\text{Level 3}$\end{tabular}}}}%
    \put(0.28077324,0.03164189){\makebox(0,0)[lt]{\lineheight{1.25}\smash{\begin{tabular}[t]{l}$\text{Level 4}$\end{tabular}}}}%
    \put(0.60639459,0.03164189){\makebox(0,0)[lt]{\lineheight{1.25}\smash{\begin{tabular}[t]{l}$\text{Level 5}$\end{tabular}}}}%
    \put(0.783778,0.03164189){\makebox(0,0)[lt]{\lineheight{1.25}\smash{\begin{tabular}[t]{l}$\text{Level 4}$\end{tabular}}}}%
  \end{picture}%
\endgroup%
}
	\caption{Illustration of a sophisticated agent designed to integrate higher-order cognitive modeling into its learning process. The agent not only learns from human interaction but also recognizes and incorporates the human's perception and understanding of the agent itself into its model. The recursive nature of this cognitive modeling, extending through multiple levels, potentially amplifies the agent's ability to ask better questions and learn efficiently from human feedback.}
	\label{fig:fig1}
\end{figure}

In stark contrast, well established models from cognitive science \cite{Stahl1995level-k, goodman2016rsa} indicate that especially in social settings, humans may have multiple levels of such situational awareness. These models not only account for how humans model other agents they interact with, but also \textit{how they can attribute such capabilities to other agents} in a recursive fashion. Furthermore, besides these models predicting human behavior more accurately in specific contexts, efficient communication, such as that entailed by the rational speech acts (RSA) framework \cite{goodman2016rsa}, \textit{might also require} higher order recursive modeling. This highlights an opportunity for improving the sample efficiency of AI-human interaction by incorporating such models as a part of AI systems. 

Moving towards the potential use of higher-order cognitive models in active learning, we present a taxonomy of interaction involving different levels of cognition (Fig.~\ref{fig:fig1}), incorporating qualitative descriptions of the types of behaviors that different levels allow for. We also provide an example of a higher-order cognitive model and demonstrate the types of behavior it can yield. Lastly, we discuss how these examples guide the conduct of future user studies and illuminate other promising research directions.

\section{Background and related work}

\subsection{Higher-order cognitive models in computational rationality}

We base our proposed framework for higher-order active learning on the theory of computational rationality, which considers human behavior to be optimal with respect to certain preferences and under constraints imposed by certain cognitive bounds \cite{gerchman2015computationalrationality}. One of the first empirically established formulations of hierarchical cognition in computational rationality is given by the level-k family of models \cite{Stahl1995level-k}, where the computational bound specifies the depth of recursion that an agent can perform. RSA \cite{goodman2016rsa} is another seminal theory based on higher-order models, which formalises pragmatic communication in terms of recursive Bayesian reasoning. 

It has been shown that similar kinds of pragmatic behavior as predicted by these models can arise in AI systems as solutions to certain types of optimisation problems \cite{malik2018efficientbellman, Fisac2020Pragmatic-pedagogic}. This is especially the case in optimal solutions to problems formalised as cooperative inverse reinforcement learning \cite{Hadfield-Menell2016cirl}, where there is an asymmetry in information that two agents possess about a cooperative task, thus requiring them to share relevant information in an efficient manner. These model-free approaches speak to the potential usefulness of higher-order interaction between humans and AI systems, but they do not directly allow for the incorporation of user models that encode more specific assumptions about human cognition. 

The explicit incorporation of cognitive science models for recursive reasoning has been demonstrated to improve the efficiency of both machine-machine interaction \cite{wen2018probabilistic, wu2021bayesiandelegation, Moreno2021NeuralRB} and human-machine interaction \cite{Ho2016showingversusdoing, Milli2019LiteralOP, Sumers2022howtotalk}. An alternative, yet more specific viewpoint on recursive cognition is the ability to attribute mental states to others in order to explain their behavior, often referred to as Theory of Mind (ToM). Inductive biases in AI systems inspired by ToM have been shown to facilitate inverse modeling of goal-driven behavior and multi-agent planning under uncertainty \cite{rabinowitz2018mtom, foerster2019bad, wu2021bayesiandelegation}. Further motivating our work, these classes of methods have recently been identified as a promising research direction for constructing the next generation of user models \cite{celikok2023modeling}. However, none of these works specifically consider the setting where the AI system learns actively from human feedback rather than passively. To the best of our knowledge, the potential utility of higher-order cognitive user models in an active learning setting has not been explored previously.

\subsection{Active learning with human feedback}

Learning actively from human feedback is an approach to machine learning where the learning system can influence what type of data the human provides. On an abstract level, this interaction happens such that the learning system asks \textit{questions}, to which the human provides \textit{answers}. 

Active learning has been shown to improve the data efficiency of learning from expert feedback, both in computational and user studies and for multiple modalities. One such modality is active learning from demonstrations (LfD), where the system queries for a demonstration from a selected starting state \cite{Silver2012ActiveLfD, Chen2020ActiveDQN}. These approaches use the decision-making uncertainty as a heuristic for deciding when it is beneficial to query for expert demonstrations. Another modality is learning from preference queries, where an expert is provided multiple options and provides the best alternative as an answer. Compared to demonstration queries, answers to preference queries in general contain less information but have also been shown to be easier for humans to provide answers to \cite{biyik2022dempref, biyik2023ActivePreferenceGP}. Feature queries are yet a third modality where active learning has been shown to improve learning efficiency \cite{Basu2018learning}. Such queries resemble preference queries, but they also allow humans to communicate in their answers the most relevant feature. These works exemplify the diversity of settings involving expert feedback where active learning can be useful. Although the case study we present builds on top of an active learning algorithm that generates preference queries that maximise the expected information gain, similarly as in \cite{biyik2022dempref}, the main difference between these works and ours is that they do not consider the expert to have a model of the system they are teaching.

Sharing certain similarities with our work is an intriguing result by \cite{colella2020strategic}, which shows that humans can strategically steer a Bayesian optimisation active learning system for improved learning performance. Unlike our work, the authors do not propose a model for how the participant performs this task. In their study, all meaningful internal states of the active learning system are also fully transparent to the participant. This raises the question whether humans can infer the internal states of an active learning system that lacks full transparency or explainability, like a system with a belief that is not easy for humans to visualise or understand. Our work, however, does not make this assumption, instead it outlines a vision where explainability emerges as a result of higher-order interactions.

\section{Levels of agency in active learning}

Here we outline our vision for higher-order active learning, which combines the paradigms of higher-order cognitive modeling and active learning from human feedback.

We consider a setting where an active learning system or artificial agent $\pi_a(\xi | b_a)$ asks questions $\xi \in \Xi$ based on its belief $b_a \in \mathcal{B}_a$ about a human teacher, with the objective of learning a model $p(y | \xi, \theta)$ based from the answers. The human provides these answers correspondingly as $y \sim \pi_{h}(y|\xi, \theta, b_h)$, based on its level 1 model $\theta \in \Theta$ and belief $b_h \in \mathcal{B}_h$ about the artificial agent. The question, for example, could be a starting state where the human provides a demonstration trajectory, or a set of items to which the human provides a ranking. The agent's and human's belief spaces $\mathcal{B}_a$ and $\mathcal{B}_h$ may be of higher order; for example a second-order belief space consists of beliefs-of-beliefs and so on. We begin with active learning involving a level 1 human and progressively describe how each level of agency up to level 5 unlocks new opportunities for enhancing AI-human interaction.

\subsection{Levels 1 and 2: Literal active learning}

We consider a level 1 human as invariant to the internal states of the agent, thus always giving answers consistent with their model $\theta$:
\begin{equation}
    \pi^{level 1}_h(y | \xi, \theta, b_h) = p(y | \xi, \theta).
\end{equation}
An active learning system, which adapts its questions based on its belief $b_a = p(\theta)$ about the level 1 user model $\theta$, is classified as a level 2 agent. Such an agent performs a posterior update based on assumingly literal answers provided by the human by
\begin{equation}
    p(\theta | \xi, y, b_a) = \frac{p(y | \xi, \theta)b_a}{p(y|\xi)}.
\end{equation}
The updated posterior belief is then used to generate the next question by maximising an objective given by
\begin{equation}
    p(\xi|b_a) \propto \exp \left[ \beta_a \mathbb{E}_{p(y|\xi,\theta)p(\theta)}\left[U^{\text{L2}}_a(y | \xi, b_a)\right] \right],
    \label{eq:level2_objective}
\end{equation}
where $\beta_a$ is a rationality coefficient for the questions and $U^{\text{L2}}_a(\cdot|\cdot, \cdot)$ is used to denote a general level 2 utility function, which we denote by uppercase L2. This utility quantifies the value of an answer with respect to the current belief, for instance, by measuring the amount of relevant information the answer provides about a specific task. Since the answer $y$ contains both aleatoric and epistemic uncertainties, the agent must optimise the expected utility considering these uncertainties. Special cases of such a level 2 utility that have been utilised in active learning from human feedback are EIG \cite{biyik2023ActivePreferenceGP} and volume removal \cite{Basu2018learning}. Importantly, level 2 active learning views questions solely  as a means to extract relevant information from the expected answer.

\subsection{Level 3: Theory of Mind reasoning and strategic teaching}

A level 3 human provides answers conditioned on a second-order nested belief $b_h = p(b_a)$, thus affording it the capacity to incorporate a model about the agent's model about the human. This in turn allows the human to make inferences about the knowledge that an active learning system has about itself based on the questions it is asking:
\begin{equation}
    p(b_a | \xi, b_h) = \frac{p(\xi | b_a)b_h}{p(\xi)}.
    \label{eq:level3tom}
\end{equation}

Modeling the human as a level 3 agent also allows for the utilisation of such beliefs obtained through ToM for shaping the belief update of the learner, referred to as strategic teaching \cite{peltola2019machine}. We consider a strategic teacher as selecting answers that shape the belief of the learner such that the probability of the user's true model $\theta$ is maximised through the following objective:
\begin{align}
    &p(y | \xi, \theta^{\text{true}}, b_h) \propto \exp \left[ \beta_h \mathbb{E}_{p(b_a)}U^{\text{L3}}_h(y | \theta^{\text{true}}, \xi, b_a)\right] \\
    &U^{\text{L3}}_h(y | \theta^{\text{true}}, \xi, b_a) = p(\theta^{\text{true}} | \xi, y, b_a) \label{eq:L3_objective},
\end{align}
where $\beta_h$ controls the rationality of the teacher.
Since the teacher has uncertainty about the learners true belief $b_a$, an optimal level 3 teacher optimises the expected utility under their second-order belief.

\subsection{Level 4: Pragmatic questions}

An active learning system that models a human as level 3 and performs reasoning in this model to generate questions is classified as a level 4 agent. As such a level 4 active learning system can use ToM to understand the intent behind the human's responses. Moreover, active learning on level 4 adds another layer of meaning to questions; while a level 2 agent uses questions as a means for receiving information, a level 4 agent may also use questions to \textit{convey} information. A similar strategic teaching formulation from the previous section yields questions that aim to make the belief of the agent \textit{identifiable} to the human:
\begin{equation}
    U^{\text{L4}}_a(\xi | b^{\text{true}}_a, b_h) = p(b^{\text{true}}_a| \xi, b_h).
\end{equation}
Evaluating this utility for a particular question requires the agent to simulate the level 3 ToM update given by Eq. \ref{eq:level3tom}, which is performed by the human after observing that question and then evaluating the resulting posterior density for its true belief $b_a^{\text{true}}$. Alternatively, a level 4 utility may also be used for communicating the relevance of a particular feature or model parameter through the question.

A weighted combination of level 2 and level 4 utilities results in an objective that optimises a tradeoff in the \textit{bidirectional information flow} between the teacher and student. 

\subsection{Level 5: Pragmatic inference}

A level 5 human can be aware of the dual nature of questions described above, therefore giving the ability of understanding the \textit{intention} behind a question. By accounting for the fact that questions might be chosen strategically, pragmatic inference from questions can for example allow for identifying what the agent considers relevant. 

The possible directions of information flow can also serve as two alternative hypotheses to explain the intentions behind questions. Reasoning about the intention in such a way can be done by computing the following Bayes factor:

\begin{equation}
    BF = \frac{\int\pi_a^{L2}(\xi | b_a)p(b_a)db_a}{\int\pi_a^{L4}(\xi | b_a)p(b_a)db_a}.
\end{equation}
To compute this Bayes factor, the human needs to marginalise over the agent's possible beliefs $b_a$. A value of $BF>1$ supports the hypothesis that the question is \textit{literal} in nature, aiming to extract information from the answer, whereas $BF<1$ supports the alternative hypothesis that the question is \textit{rhetorical} in nature, aiming to convey information with the question. For the case where the human attributes a level 2 model to the agent that is based on EIG, an intuitive rhetorical question (e.g. "is this not how it is?") is a question which has an obvious answer for every possible agent belief $b_a$, resulting in a low marginal likelihood $\int\pi_a^{L2}(\xi | b_a)p(b_a)db_a$, therefore likely also resulting in $BF << 1$. Taking this example one step further, a level 6 agent that wishes to communicate intention through questions might strategically choose a question with an obvious answer in order to make the rhetorical nature of the question identifiable to the human.

\section{Case study: Higher-order active learning with preference queries}

We now move to a computational study of active learning with preference queries, illustrating how the general methods depicted in the previous section can be grounded into concrete models and how these models can influence the  resulting interactions. We have designed experiments for a computational model of a level 3 human, aiming to answer the following questions:

\begin{enumerate}
    \item Can a level 3 human model infer the belief of a level 2 active learning system by observing queries aimed at reducing uncertainty about a unimodal belief?
    \item Is it possible to identify such a belief as bimodal by observing queries that aim to reduce uncertainty both about \textit{the group} the human belongs to and the parameters \textit{within the group}?
    \item How does belief attribution affect level 3 strategic teaching behavior? In particular, can beliefs inferred by ToM from preference queries lead to qualitative changes in the teaching strategy?
\end{enumerate}

\subsection{Strategic teaching scenario}

To ground our case study with a concrete example, consider a recommender system for a car shopping website that has the objective of identifying the price which best aligns with the preferences of the user. The system has a bimodal prior belief for the desired price, encoding domain knowledge about the presence of two distinct user groups regularly visiting the site: one group is looking for cheaper used cars and the other group for more luxurious ones. The system offers two options at a time to the user from which they are supposed to select the more suitable one. If the user makes certain assumptions about how recommendations are produced, such as their purpose being to gather information from the answer, they may employ ToM reasoning to identify internal beliefs of the system. 

\subsubsection{Human and Agent Models:}

We next proceed to a more detailed description of a computational model that can allow for such reasoning in our example scenario. This model describes a level 3 human, who itself models the agent it is interacting with as level 2. As such, the human also has a model of the level 1 model that the level 2 agent has about its own preferences.

We start by describing this level 1 model, which defines how humans respond to preference queries. In this model answers are consistent with the true preference of the human as follows:

\begin{equation}
    p(y=1 | \theta, x_1, x_2) = \sigma(r_{\theta}(x_2) - r_{\theta}(x_1)),
\end{equation}
where $r_{\theta} : \mathbb{R} \rightarrow \mathbb{R}$ is a stochastic reward function which depends on the distance to the most preferred item $\theta$, $\sigma : \mathbb{R} \rightarrow [0, 1]$ is the logistic sigmoid function, $x_1, x_2 \in \mathbb{R}$ are 1-dimensional real-valued item features (normalised prices in our example) and $y \in \{0,1\}$ is binary answer where $y=1$ implies that $x_2$ is preferred over $x_1$. We model a setting where there are two distinct groups of human preferences with the following bimodal prior for $\theta$ as

\begin{align}
    &p(z = 1) = p_z \\
    &p(\theta_1) =  \mathcal{N}(\theta_1 | \mu_1, \sigma_1^2) \\
    &p(\theta_2) =  \mathcal{N}(\theta_2 | \mu_2, \sigma_2^2) \\
    &\theta = z \theta_1 + (1 - z) \theta_2,
\end{align}
where $\mu_i$ and $\sigma^2_i$ are the mean and variance of the preferences of group $i$ and $z \in \{0, 1\}$ is a binary latent variable that identifies the group.

The objective of the active learning system is to maximise the expected utility of the answers as given by (\ref{eq:level2_objective}). A natural choice for such a local utility is the information gain, leading to a level 2 agent that maximises the mutual information between its model and the answers as

\begin{align}
    &p(x_1, x_2 | b_a) \propto \exp(\beta_a I(\theta, y | x_1, x_2, b_a)) \\
    &I(\theta, y | x_1, x_2, b_a) = \mathbb{E}_{p(y|x_1,x_2)}\left[U^{\text{IG}}(y| x_1, x_2, b_a)\right] \\
    &U^{\text{IG}}(y| x_1, x_2, b_a) = H(p(\theta)) - H(p(\theta | x_1, x_2, y)),
\end{align}
where $H(\cdot)$ is the information entropy.

After observing a set of one or multiple preference queries, $\mathcal{D}_x=\{(x_1^i, x_2^i)\}_{i=1}^N$, we estimate the agent's belief $b_a$ with maximum likelihood as follows:

% \begin{align}
%     &p(b_a | \mathcal{D}_x) = \frac{\prod_{i=1}^Np(x_1^i, x_2^i|b_a)p(b_a)}{p(\mathcal{D}_x)} \\
%     &\propto \exp\left[\beta_a \sum_{i=1}^N\left[I(\theta, y | x_1, x_2, b_a)\right]\right]p(b_a).
% \end{align}

\begin{align}
    \hat{b}_a &= \argmax_{b_a}p(\mathcal{D}_x | b_a) \\
    &=\argmax_{b_a}\prod_{i=1}^N\exp\left[\beta_a I(\theta, y | x^i_1, x^i_2, b_a)\right] \\
    &=\argmax_{b_a}\sum_{i=1}^NI(\theta, y | x^i_1, x^i_2, b_a).
\end{align}
Thus, in this model the human explains the observed questions by attributing to the agent a belief that makes those questions maximally informative.

\subsection{Identifiability of unimodal queries}

We first study how well the belief of the active learning system can be identified by our user model when queries aim to reduce uncertainty about one group only. To create such a setting, we use a prior where $\mu_1=-3, \sigma_1^2=1, \mu_2 = 3, \sigma_2^2=1$ and $p_z=0.9$. 

\begin{figure}[t]
\centering
\includegraphics[width=1.0\columnwidth]{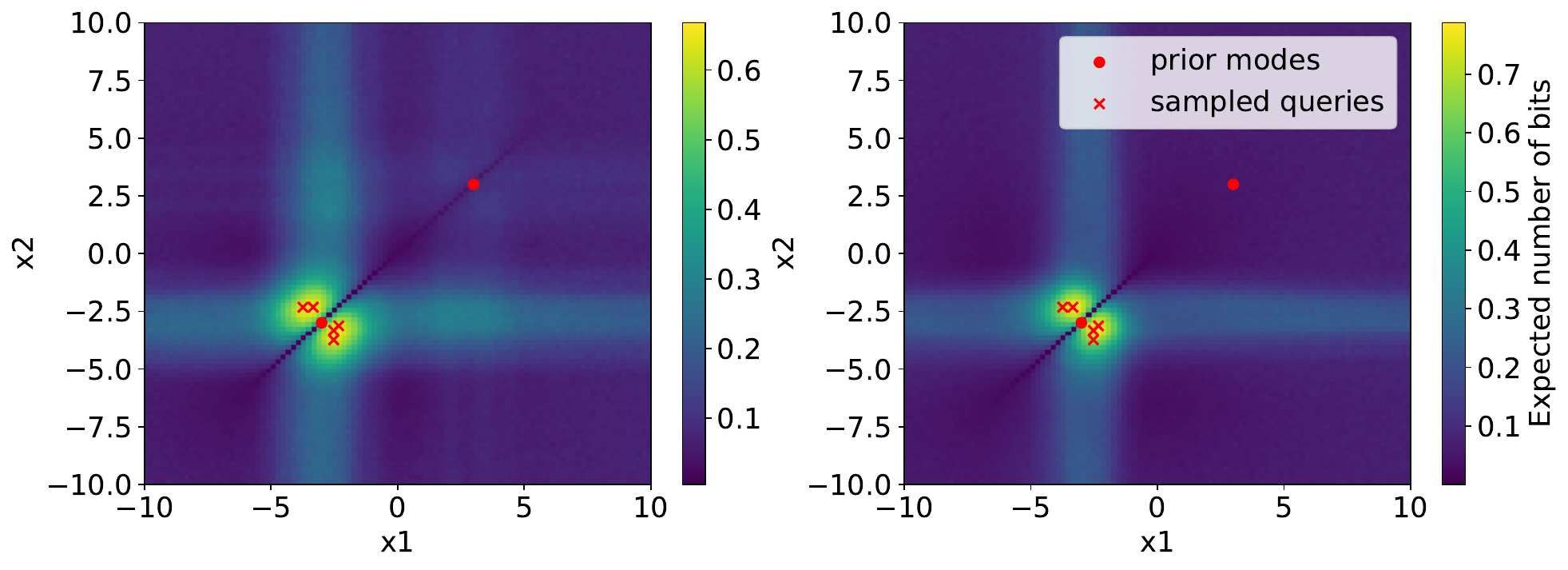}
\caption{Identifiability of agent belief $b_a$ from unimodal preference queries. Left: EIG visualised in query space for the true belief. Right: EIG prediction for the belief obtained with maximum likelihood from the 5 queries shown.}
\label{fig:unimodal_eig}
\end{figure}

Figure \ref{fig:unimodal_eig} shows sampled queries and the EIG prediction in the query space for the belief given by the maximum likelihood estimate. This example shows how it from preference queries is possible to \textit{identify the belief that was used to generate the queries}. The estimated belief allows for making predictions for the EIG that resembles those obtained from the true belief.

\subsection{Identifiability of bimodal queries}

Next we demonstrate how identifiability is affected by the uncertainty regarding the group, resulting in queries that aim to reduce uncertainty about both the group and the modes. For this setting, we use the same parameters for the group means, but set $\sigma_{1,2}=0.5$ and $p_z = 0.6$, which keeps the total uncertainty similar but spreads it over both modes. We employ $N=20$ queries here since the inference problem is more difficult in the bimodal case.

\begin{figure}[t]
\centering
\includegraphics[width=1.0\columnwidth]{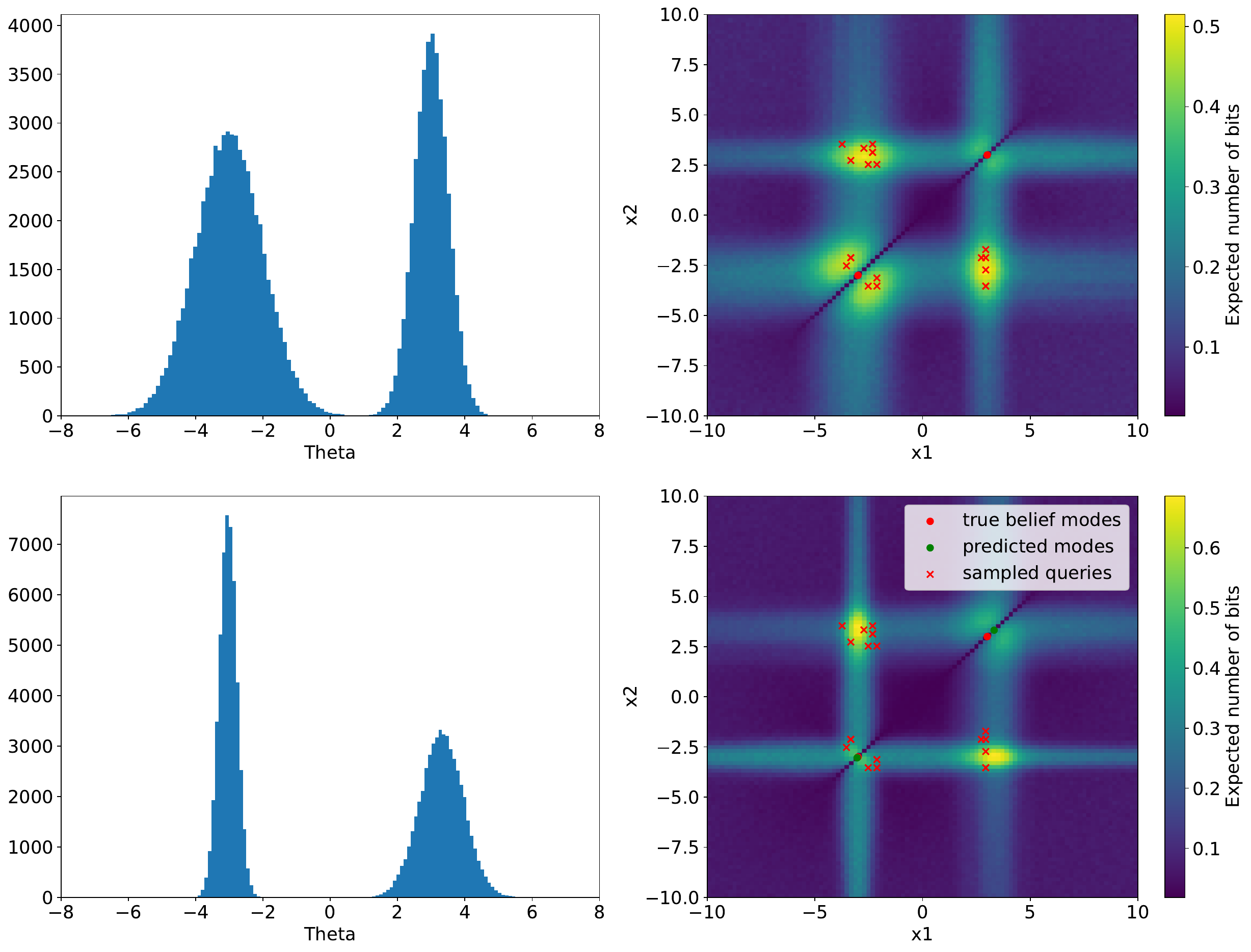}
\caption{Identifiability of agent belief $b_a$ from bimodal preference queries. Top left: Samples from the true belief. Top right: EIG and queries resulting from true belief. Bottom left: Belief estimated with maximum likelihood from sampled queries. Bottom right: EIG prediction obtained from estimated belief.}
\label{fig:bimodal_eig}
\end{figure}

An example outcome of such inference is shown in Figure \ref{fig:bimodal_eig}. This example shows that although the exact belief may not always be accurately identified, it remains possible to detect the \textit{locations of the modes} and determine if there is \textit{uncertainty regarding the group} to which the participant belongs.

\subsection{Effect of belief attribution on strategic teaching}

Lastly, we show how our model of a level 3 human can detect a \textit{false belief} about itself through preference queries generated by an active learning system and use strategic teaching to correct it. We use a similar prior and dataset to that in the unimodal queries example (Figure \ref{fig:unimodal_eig}), with the user model parameter $\theta^{\text{true}}=2.0$ representing the ground truth, which belongs to the less likely mode. To illustrate the impact of the belief derived from such inference on teaching behavior, we employ a similar type of strategic teaching objective as given in (\ref{eq:L3_objective}). However, here it is applied to selecting full data points $(x_1, x_2, y)$ rather than the answers only.

\begin{figure}[t]
\centering
\includegraphics[width=1.0\columnwidth]{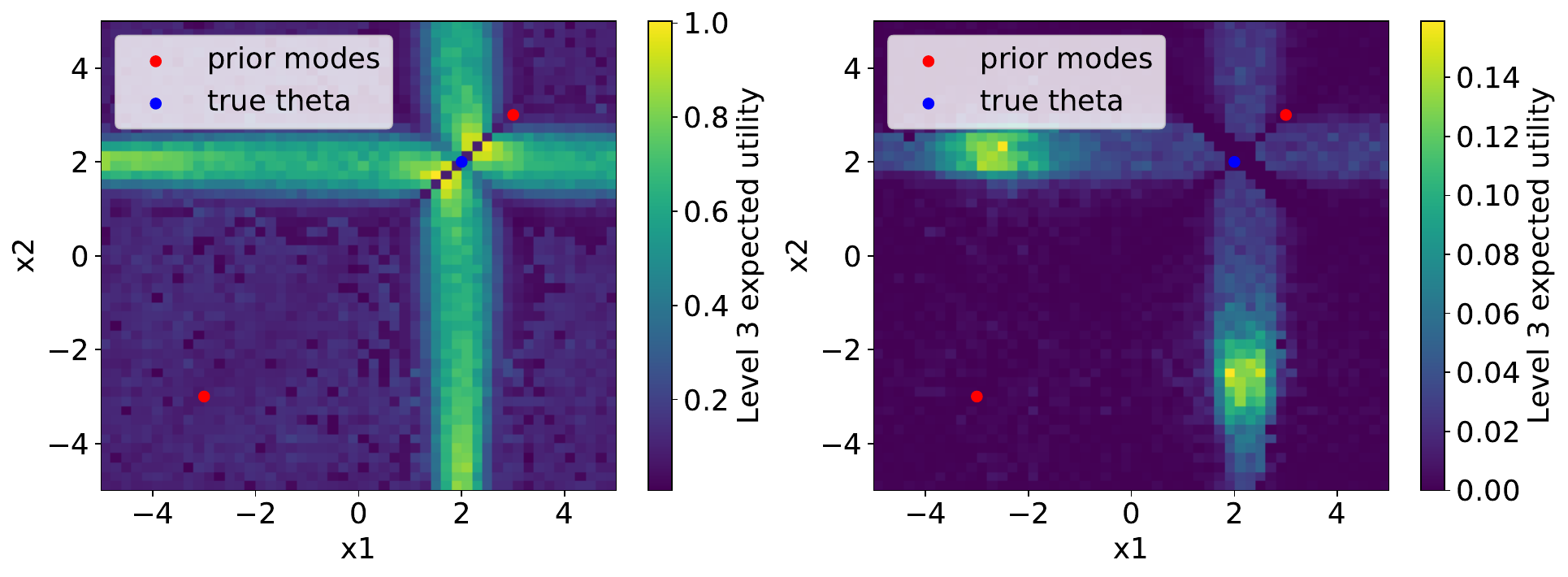}
\caption{An example of how the identification of a false belief can impact strategic teaching behavior. Left: Expected utilities of teaching data points $x_1, x_2$ for a uniform belief $b_a$. Right: Corresponding expected utilities for belief inferred after observing 5 preference queries that were generated from the false belief.}
\label{fig:belief_correction}
\end{figure}

Figure \ref{fig:belief_correction} presents an example of a qualitative difference that can arise between a level 3 human that does not adapt their belief about the agent's knowledge based on the observed queries, and one who integrates ToM inference with strategic teaching. In this illustration, the non-adaptive model shows a teaching example where both options closely align with the true preference. Conversely, \textit{the adaptive model shows an example that compares the mode of the false belief with the true preference}. Integrating information about both the false belief and truth in the teaching example is an intuitive strategy for steering the agent's belief towards the correct mode.

\section{Discussion}

In our computational study, we demonstrated examples of inferences performed by a level 3 human model based on the queries from an active learning system, along with how these inferences can affect behavior. The demonstrated behavior serves as a motivating example for how a level 3 human could recognise internal states of an active learning agent, such as uncertainties or false beliefs, and then adapt their behavior accordingly to enhance the effectiveness of their teaching. The advantage of the demonstrated adaptation is also intuitive; a strategic teaching approach that incorporates information about both the false belief and the ground truth is likely to be more efficient.

The potential demonstrated by our computational user model encourages further computational studies. Our case study considers a particular instance of the class of abstract level 3 user models that we have described, wherein true human preferences are assumed to follow a simple logistic model, and where humans assume that questions are generated with the objective of maximizing their EIG. Characterizing the extent to which various modeling assumptions influence the resultant behavior would be beneficial. For example, how does the identifiablity of the active learning system's belief change when queries are generated with different level 2 objectives?

Such a characterisation of the effect of modeling assumptions in a computational setting can in turn guide empirical user studies. If different computational models result in distinct behaviors, a natural next step is to empirically select the most appropriate model among them.  Additionally, our case study also informs us how the beliefs that participants of a user study attribute to active learning systems could be predicted from their behavior. Specifically, if users engage in strategic teaching behavior that specifically aims to correct false beliefs, it provides evidence that they are capable of recognising such false beliefs through ToM. Furthermore, a comparison between the strategy employed by a participant and predictions given by a computational model could allow for more specific conclusions about the beliefs that humans attribute to active learning systems, as well as the factors that affect them.

The empirical validation of a level 3 user model would justify its incorporation into a level 4 active learning agent. While the learning efficiency of a level 2 active learning agent already in principle can benefit from level 3 human behavior, the full potential of communication with a level 3 human, as described in our taxonomy, is to be reached when the agent can pragmatically query the user. However, at some stage, it is anticipated that approximations to optimal communication will become necessary, both within user models and active learning algorithms. Consequently, future research should focus on identifying the types of approximations that maintain the qualitative modes of behavior predicted by computationally rational higher-order models. This research would guide the development of active learning methods that can fully utilise the advantages of complex and nuanced human behavior.

\section{Interdisciplinary implications}

The methods we have described build on approaches from both the fields of machine learning and cognitive science. The cognitive model we describe in our case study illustrates how machine learning algorithms---in our case mutual information estimation originating from Bayesian experimental design \cite{rainforth2023modern}---may provide interesting hypotheses for user models to be tested in empirical user studies. The converse may also be true; results from the kinds of user studies we have proposed can guide the search for better machine learning methods. For example, constructing a level 5 user model that balances computational efficiency with accurate representation of human behavior might necessitate advanced algorithmic methods surpassing the state of the art. Eventually, the long-term goal of constructing AI agents that utilise the kinds of user models we advocate for, have the potential to enhance the sample efficiency, explainablity, safety and reliability of future machine learning systems.

\section{Conclusion}

In this work, we have proposed to combine higher-order cognitive models with active learning methods based on human feedback. Although previous works have explored such models in learning from human feedback, the methods we outline fill a gap in understanding how these models could specifically be applied in the active learning setting. We present theoretical arguments for the utility of such a framework in active learning with human feedback and also provide a practical example where a higher-order cognitive model recognizes and corrects a false belief in an active learning system. Building on these findings, we explore the implications for designing future computational and user studies, which represents a critical next step in applying the proposed methods effectively in real-world systems.

\bibliography{refs}

\begin{thebibliography}{23}
\providecommand{\natexlab}[1]{#1}

\bibitem[{Basu, Singhal, and Dragan(2018)}]{Basu2018learning}
Basu, C.; Singhal, M.; and Dragan, A.~D. 2018.
\newblock Learning from Richer Human Guidance: Augmenting Comparison-Based Learning with Feature Queries.
\newblock In \emph{Proceedings of the 2018 ACM/IEEE International Conference on Human-Robot Interaction}, HRI '18, 132–140. New York, NY, USA: Association for Computing Machinery.
\newblock ISBN 9781450349536.

\bibitem[{Bıyık et~al.(2023)Bıyık, Huynh, Kochenderfer, and Sadigh}]{biyik2023ActivePreferenceGP}
Bıyık, E.; Huynh, N.; Kochenderfer, M.; and Sadigh, D. 2023.
\newblock Active preference-based Gaussian process regression for reward learning and optimization.
\newblock \emph{The International Journal of Robotics Research}.

\bibitem[{Bıyık et~al.(2022)Bıyık, Losey, Palan, Landolfi, Shevchuk, and Sadigh}]{biyik2022dempref}
Bıyık, E.; Losey, D.~P.; Palan, M.; Landolfi, N.~C.; Shevchuk, G.; and Sadigh, D. 2022.
\newblock Learning reward functions from diverse sources of human feedback: Optimally integrating demonstrations and preferences.
\newblock \emph{The International Journal of Robotics Research}, 41(1): 45--67.

\bibitem[{{\c C}elikok, Murena, and Kaski(2023)}]{celikok2023modeling}
{\c C}elikok, M.~M.; Murena, P.-A.; and Kaski, S. 2023.
\newblock Modeling needs user modeling.
\newblock \emph{Frontiers in Artificial Intelligence}, 6.

\bibitem[{Chen et~al.(2020)Chen, Tangkaratt, Lin, and Sugiyama}]{Chen2020ActiveDQN}
Chen, S.-A.; Tangkaratt, V.; Lin, H.-T.; and Sugiyama, M. 2020.
\newblock Active deep Q-learning with demonstration.
\newblock \emph{Machine Learning}, 109(9): 1699--1725.

\bibitem[{Colella et~al.(2020)Colella, Daee, Jokinen, Oulasvirta, and Kaski}]{colella2020strategic}
Colella, F.; Daee, P.; Jokinen, J.; Oulasvirta, A.; and Kaski, S. 2020.
\newblock Human Strategic Steering Improves Performance of Interactive Optimization.
\newblock In \emph{Proceedings of the 28th ACM Conference on User Modeling, Adaptation and Personalization}, UMAP '20, 293–297. New York, NY, USA: Association for Computing Machinery.
\newblock ISBN 9781450368612.

\bibitem[{Fisac et~al.(2020)Fisac, Gates, Hamrick, Liu, Hadfield-Menell, Palaniappan, Malik, Sastry, Griffiths, and Dragan}]{Fisac2020Pragmatic-pedagogic}
Fisac, J.~F.; Gates, M.~A.; Hamrick, J.~B.; Liu, C.; Hadfield-Menell, D.; Palaniappan, M.; Malik, D.; Sastry, S.~S.; Griffiths, T.~L.; and Dragan, A.~D. 2020.
\newblock Pragmatic-Pedagogic Value Alignment.
\newblock In Amato, N.~M.; Hager, G.; Thomas, S.; and Torres-Torriti, M., eds., \emph{Robotics Research}, 49--57. Cham: Springer International Publishing.

\bibitem[{Foerster et~al.(2019)Foerster, Song, Hughes, Burch, Dunning, Whiteson, Botvinick, and Bowling}]{foerster2019bad}
Foerster, J.; Song, F.; Hughes, E.; Burch, N.; Dunning, I.; Whiteson, S.; Botvinick, M.; and Bowling, M. 2019.
\newblock {B}ayesian Action Decoder for Deep Multi-Agent Reinforcement Learning.
\newblock In Chaudhuri, K.; and Salakhutdinov, R., eds., \emph{Proceedings of the 36th International Conference on Machine Learning}, volume~97 of \emph{Proceedings of Machine Learning Research}, 1942--1951. PMLR.

\bibitem[{Gershman, Horvitz, and Tenenbaum(2015)}]{gerchman2015computationalrationality}
Gershman, S.~J.; Horvitz, E.~J.; and Tenenbaum, J.~B. 2015.
\newblock Computational rationality: A converging paradigm for intelligence in brains, minds, and machines.
\newblock \emph{Science}, 349(6245): 273--278.

\bibitem[{Goodman and Frank(2016)}]{goodman2016rsa}
Goodman, N.~D.; and Frank, M.~C. 2016.
\newblock Pragmatic Language Interpretation as Probabilistic Inference.
\newblock \emph{Trends in Cognitive Sciences}, 20(11): 818--829.

\bibitem[{Hadfield-Menell et~al.(2016)Hadfield-Menell, Dragan, Abbeel, and Russell}]{Hadfield-Menell2016cirl}
Hadfield-Menell, D.; Dragan, A.; Abbeel, P.; and Russell, S. 2016.
\newblock Cooperative Inverse Reinforcement Learning.
\newblock In \emph{Proceedings of the 30th International Conference on Neural Information Processing Systems}, NIPS'16, 3916–3924. Red Hook, NY, USA: Curran Associates Inc.
\newblock ISBN 9781510838819.

\bibitem[{Ho et~al.(2016)Ho, Littman, MacGlashan, Cushman, and Austerweil}]{Ho2016showingversusdoing}
Ho, M.~K.; Littman, M.; MacGlashan, J.; Cushman, F.; and Austerweil, J.~L. 2016.
\newblock Showing versus doing: Teaching by demonstration.
\newblock In Lee, D.; Sugiyama, M.; Luxburg, U.; Guyon, I.; and Garnett, R., eds., \emph{Advances in Neural Information Processing Systems}, volume~29. Curran Associates, Inc.

\bibitem[{Malik et~al.(2018)Malik, Palaniappan, Fisac, Hadfield-Menell, Russell, and Dragan}]{malik2018efficientbellman}
Malik, D.; Palaniappan, M.; Fisac, J.; Hadfield-Menell, D.; Russell, S.; and Dragan, A. 2018.
\newblock An Efficient, Generalized {B}ellman Update For Cooperative Inverse Reinforcement Learning.
\newblock In Dy, J.; and Krause, A., eds., \emph{Proceedings of the 35th International Conference on Machine Learning}, volume~80 of \emph{Proceedings of Machine Learning Research}, 3394--3402. PMLR.

\bibitem[{Milli and Dragan(2019)}]{Milli2019LiteralOP}
Milli, S.; and Dragan, A.~D. 2019.
\newblock Literal or Pedagogic Human? Analyzing Human Model Misspecification in Objective Learning.
\newblock In \emph{Conference on Uncertainty in Artificial Intelligence}.

\bibitem[{Moreno et~al.(2021)Moreno, Hughes, McKee, Pires, and Weber}]{Moreno2021NeuralRB}
Moreno, P.; Hughes, E.; McKee, K.~R.; Pires, B.~{\'A}.; and Weber, T. 2021.
\newblock Neural Recursive Belief States in Multi-Agent Reinforcement Learning.
\newblock \emph{ArXiv}, abs/2102.02274.

\bibitem[{Peltola et~al.(2019)Peltola, {\c C}elikok, Daee, and Kaski}]{peltola2019machine}
Peltola, T.; {\c C}elikok, M.; Daee, P.; and Kaski, S. 2019.
\newblock Machine Teaching of Active Sequential Learners.
\newblock In \emph{33rd Conference on Neural Information Processing Systems}, Advances in Neural Information Processing Systems, 11202--11213. United States: Neural Information Processing Systems Foundation.

\bibitem[{Rabinowitz et~al.(2018)Rabinowitz, Perbet, Song, Zhang, Eslami, and Botvinick}]{rabinowitz2018mtom}
Rabinowitz, N.; Perbet, F.; Song, F.; Zhang, C.; Eslami, S. M.~A.; and Botvinick, M. 2018.
\newblock Machine Theory of Mind.
\newblock In Dy, J.; and Krause, A., eds., \emph{Proceedings of the 35th International Conference on Machine Learning}, volume~80 of \emph{Proceedings of Machine Learning Research}, 4218--4227. PMLR.

\bibitem[{Rainforth et~al.(2023)Rainforth, Foster, Ivanova, and Smith}]{rainforth2023modern}
Rainforth, T.; Foster, A.; Ivanova, D.~R.; and Smith, F.~B. 2023.
\newblock Modern Bayesian Experimental Design.
\newblock arXiv:2302.14545.

\bibitem[{Silver, Bagnell, and Stentz(2012)}]{Silver2012ActiveLfD}
Silver, D.; Bagnell, J.~A.; and Stentz, A. 2012.
\newblock Active learning from demonstration for robust autonomous navigation.
\newblock In \emph{2012 IEEE International Conference on Robotics and Automation}, 200--207.

\bibitem[{Stahl and Wilson(1995)}]{Stahl1995level-k}
Stahl, D.~O.; and Wilson, P.~W. 1995.
\newblock On Players' Models of Other Players: Theory and Experimental Evidence.
\newblock \emph{Games and Economic Behavior}, 10(1): 218--254.

\bibitem[{Sumers et~al.(2022)Sumers, Hawkins, Ho, Griffiths, and Hadfield-Menell}]{Sumers2022howtotalk}
Sumers, T.; Hawkins, R.; Ho, M.~K.; Griffiths, T.; and Hadfield-Menell, D. 2022.
\newblock How to talk so AI will learn: Instructions, descriptions, and autonomy.
\newblock In Koyejo, S.; Mohamed, S.; Agarwal, A.; Belgrave, D.; Cho, K.; and Oh, A., eds., \emph{Advances in Neural Information Processing Systems}, volume~35, 34762--34775. Curran Associates, Inc.

\bibitem[{Wen et~al.(2019)Wen, Yang, Luo, Wang, and Pan}]{wen2018probabilistic}
Wen, Y.; Yang, Y.; Luo, R.; Wang, J.; and Pan, W. 2019.
\newblock Probabilistic Recursive Reasoning for Multi-Agent Reinforcement Learning.
\newblock In \emph{International Conference on Learning Representations}.

\bibitem[{Wu et~al.(2021)Wu, Wang, Evans, Tenenbaum, Parkes, and Kleiman-Weiner}]{wu2021bayesiandelegation}
Wu, S.~A.; Wang, R.~E.; Evans, J.~A.; Tenenbaum, J.~B.; Parkes, D.~C.; and Kleiman-Weiner, M. 2021.
\newblock Too Many Cooks: Bayesian Inference for Coordinating Multi-Agent Collaboration.
\newblock \emph{Topics in Cognitive Science}, 13(2): 414--432.

\end{thebibliography}

\end{document}